\documentclass{article}

\usepackage{PRIMEarxiv}

\usepackage[utf8]{inputenc} 
\usepackage[T1]{fontenc}    
\usepackage{hyperref}       
\usepackage{url}            
\usepackage{booktabs}       
\usepackage{amsfonts}       
\usepackage{nicefrac}       
\usepackage{microtype}      
\usepackage{lipsum}
\usepackage[numbers]{natbib}
\usepackage{fancyhdr}       
\usepackage{graphicx}       
\graphicspath{{media/}}     

\title{PyCellMech: A shape-based feature extraction pipeline for use in medical and biological studies}

\author{
  Janan Arslan, Henri Chhoa, Ines Khemir, Romain Valabregue\\
  Sorbonne Université, Institut du Cerveau\\
  Paris Brain Institute\\
  ICM, CNRS, Inria, Inserm\\
  AP-HP, Hôpital de la Pitié Salpêtrière
  F-75013\\
  Paris, France\\
  \texttt{\{janan.arslan, henri.chhoa, ines.khemir, romain.valabregue\}@icm-institute.org} \\
   \And
  Kurt K. Benke \\
  School of Engineering \\
  University of Melbourne \\
  Parkville, Victoria, Australia\\
  \texttt{kbenke@unimelb.edu.au} \\
}

\begin{document}
\maketitle

\footnotetext[1]{Preprint. Under Review.}
\begin{abstract}
\textbf{Summary:} Medical researchers obtain knowledge about the prevention and treatment of disability and disease using physical measurements and image data. To assist in this endeavor, feature extraction packages are available that are designed to collect data from the image structure. In this study, we aim to augment current works by adding to the current mix of shape-based features. The significance of shape-based features has been explored extensively in research for several decades, but there is no single package available in which all shape-related features can be extracted easily by the researcher. \emph{PyCellMech} has been crafted to address this gap. The \emph{PyCellMech} package extracts three classes of shape features, which are classified as one-dimensional, geometric, and polygonal. Future iterations will be expanded to include other feature classes, such as scale-space. 
\newline\textbf{Availability and implementation:} PyCellMech is freely available at \url{https://github.com/icm-dac/pycellmech}.

\end{abstract}


\section{Introduction}
Medical research involves understanding the drivers behind disease prevalence and progression. Typically, such research is investigated using a combination of statistical or artificial intelligence (AI)-driven methods that could involve mining text and imaging data. Other evaluations include using both these media to extract potential features that may be characteristic of a disease state. For example, in the comprehensive and popular work of van Griethuysen \emph{et al.} (2017), researchers can extract a large list of image-related features (typically for radiological images) using the Python-based package \emph{pyradiomics} (\cite{vanGriethuysen2017}). This list includes first-order statistics, along with texture statistics, such as gray-level co-occurrence matrix (GLCM). 

Amongst the existing body of work, another consideration can be found in the analysis of shape-based or morphometric features. This field of study first arose to prominence in the 1980s and investigated the incorporation of biological shape information. This includes translating shape descriptions in the form of quantitative measures that do not vary even when the shape is moved, rotated, enlarged, or reduced (\cite{Bookstein1997}). Similar to existing packages, shape-related information can be used to help derive similarities or differences in disease classes, and thus provide additional insights regarding disease pathology and progression. In the literature, some packages already exist and take into consideration some shape-based features. These include perimeter, area, and aspect ratio (\cite{Lobo2015}), as well as others, such as mesh surface and surface ratio currently offered by \emph{pyradiomics}. Additionally, other feature extraction works include the \emph{geomorph} package in R (\cite{Adams2013}), which is used to evaluate anatomical features of organisms for ecological and evolutionary studies. 

In this paper, we augment this body of work by developing an easily accessible, open-source package that includes additional shape features not yet considered. Our objective is to provide complementary features that researchers can use when elucidating disease mechanics. Our package, named \emph{PyCellMech}, is launched with a preliminary set of shape-based features inspired by the works of \cite{ChakiDey2019} and \cite{Yin2008}. To the best of our knowledge, these features have not been made collectively available to the scientific community. In this paper, we discuss the current iteration of the pipeline, present results based on open-source data, and conclude with future directions.

\section{Approach}\label{sec2}

\begin{figure*}[!t]%
\centering
  \caption{Overview of \emph{PyCellMech} pipeline}
  \includegraphics[width=0.8\linewidth]{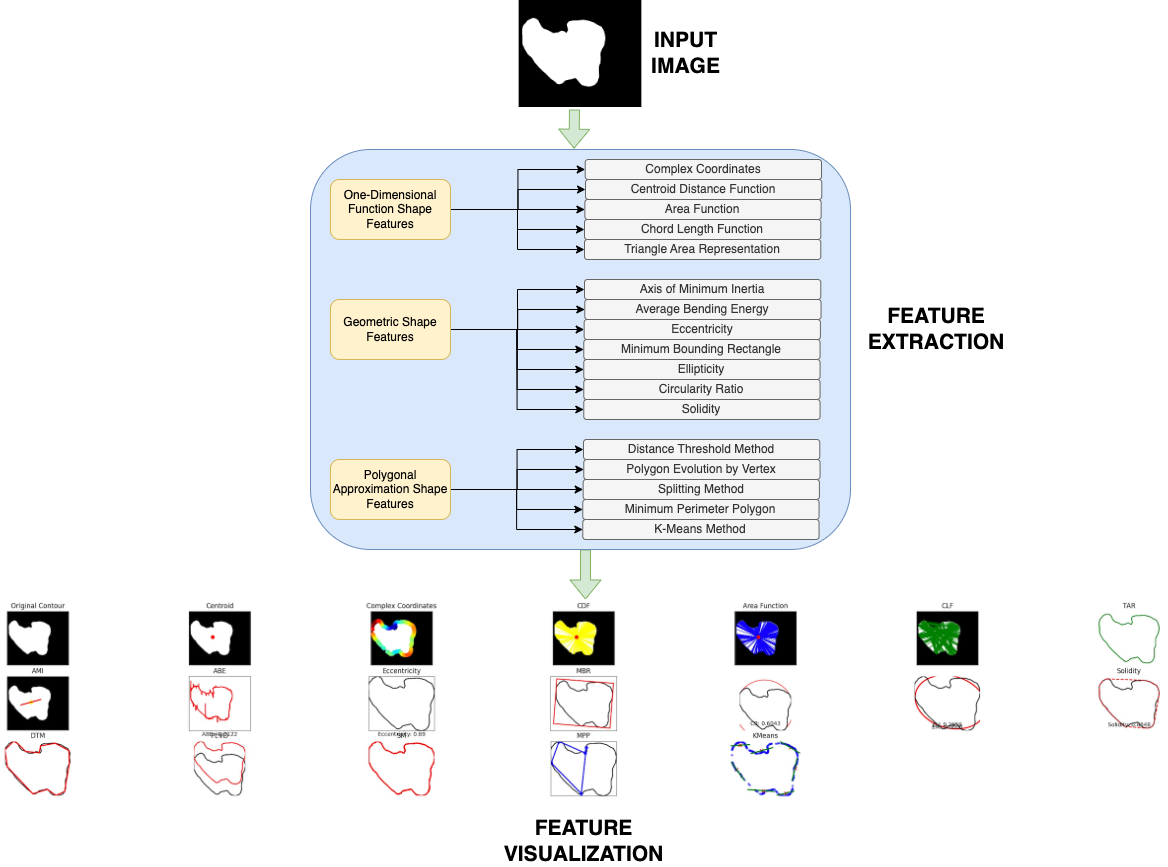}
\label{fig:fig1}
\end{figure*}

The framework for the \emph{PyCellMech} pipeline is outlined in Fig~\ref{fig:fig1}. The pipeline can be installed in Python using the command \verb|pip install pycellmech|. 

The pipeline consists of three primary components: image pre-processing, feature extraction, and feature visualization. For its input, the pipeline currently takes binarized images; this is in contrast to existing packages, such as \emph{pyradiomics}, which read the binary masks in parallel to the original image for feature extraction (e.g., extracting texture or intensity-related information). However, the sole focus of this pipeline is understanding the inherent shape properties of the regions of interest (ROIs), and such information is sufficiently attainable in the binarized marks. The pipeline can process both single- or multi-class information (refer to Section \ref{sec3} for further details).

The entire pipeline can be run using the command line illustrated below:
\begin{verbatim}
    pycellmech 
    --input /path/to/folder/with/binary/masks
    --csv_file /path/to/save/extracted/features/as/csv 
    --output /path/to/saving/feature/maps
    --label 's' (single-class) / 'm' (multi-class)
    --nifti_folder /path/to/muti-class/nifti/metadata
\end{verbatim}

An end-user can save their collection of binarized masks into a single folder, which can be called upon from the command line. All traditional image formats are accepted as input, including JPEG, PNG, and TIF. These images undergo a pre-processing of Gaussian blurring and morphological closing. The pipeline then iterates through every image within the folder, identifying all the ROIs within each image. All shape features are calculated and processed for each ROI; this information is output as a CSV file. An illustrative example of the CSV output can be found in Table~\ref{tab:tab1}. For ease of accessibility, features extracted are formatted in the following manner: image name, the \emph{i}th ROI processed in the image, its label (if multi-class is selected), and all the shape features as related to the \emph{i}th ROI. For example, in Table~\ref{tab:tab1}, image \verb|ck2bxskgxxzfv08386xkqtqdy.jpg| had two ROIs processed. 

Parallel to the quantification of the shape features, we also offer a qualitative assessment through the visualization of the collected shape features (we refer to this visualization as a feature map). In this feature map, the largest ROI based on connected regions in each image is automatically extracted. The original binary mask is illustrated as a reference image. All shape features are iteratively displayed in the feature map via superimposing each feature on the largest ROI and displaying it accordingly (refer to \emph{Feature Visualization} in Figure~\ref{fig:fig1}). This task is repeated and saved for every binary mask within the chosen folder. These feature maps are saved using the argument \verb|--output| in the command line.

In its inaugural version, the pipeline consists of three feature classes: one-dimensional function, geometric, and polygonal approximation. The choice of these modules was simply due to their practicality, given their ease of implementation and larger potential spectrum of application. One-dimensional function features hold the perspective properties of the shape (e.g., the Centroid Distance Function [CDF] estimates the distance between the center of an object to its boundary, describing the shape's geometry relative to its center). Geometric shape features are used to determine the likeness amongst shape characteristics (e.g., Average Bending Energy [ABE] quantifies the mean effort required to deform a shape). Polygonal approximations overlook edge discrepancies (e.g., Minimum Perimeter Polygon [MPP] refers to a polygon with the smallest possible perimeter that can enclose a given shape). Further details regarding each feature can be found in the \emph{PyCellMech} GitHub Repository. These modules will continually be built upon in future iterations of the pipeline as new research in the space becomes available.

The design and preparation of this pipeline were done using the open-source segmented polyp dataset, Kvasir, available through Kaggle (\cite{KvasirDataset}) as well as internal data available to our team not yet published. 

\section{Single vs. Multi-Class Feature Extraction}\label{sec3}
Users have the option of selecting 's' for images with a single class or 'm' for multi-class. Multi-class, for example, could represent the presence of multiple cell types within each image. This functionality allows the users greater flexibility, being able to automatically assign pre-defined labels alongside extracted shape features. If users opt for the single-class option, no labels need to accompany the binarized images. For users working with multi-classed masks, they are required to have labels in NifTI (Neuroimaging Informatics Technology Initiative) format which stores labeled metadata. NifTI files are preferable as they offer the following advantages: (1) they preserve spatial context information, (2) improve data organization, and (3) can easily be validated (e.g., visualizing the original image and labels). If users do not have at their disposal pre-existing NifTI files, \emph{PyCellMech} accommodates for this possibility, allowing the users to create one from scratch using the \verb|pycellmech_create_label| and \verb|pycellmech_nifti| 
 functions in the package accordingly. The following steps are recommended to prepare images for multi-class processing:

\begin{itemize}
    \item Using the \verb|pycellmech_create_label| command line, specify the folder in which the original binarized images are stored. For each image in this folder, the function will label each ROI based on connected regions, returning a numerical label for each ROI within the image.
    
\begin{verbatim}
    pycellmech_create_label 
    --folder_path /path/to/folder/with/binary/masks
    --output_csv_folder /path/to/save/labels/as/csv 
    --output_image_folder /path/to/save/labeled/masks
\end{verbatim}
    
    \item Outputs from this function will include (1) the binarized image converted to an RGB format with the labels superimposed for visualization, and (2) a CSV file that contains the headers \verb|cell_id| (with this representing the ROI labels generated using connected regions), \verb|centroid_x|, and \verb|centroid_y| that represent the centroid coordinates of each ROI detected, and an empty column named \verb|final_label|. 
    \item The end-user will need to fill in the \verb|final_label| column. The RGB image with the \verb|cell_id| superimposed on each region can be used as a reference when annotating the data. We strongly recommend the use of numerical values to classify the ROI. For example, to classify the level of differentiation in cells, users can input 0 for undefined cells (which will be converted to the background during feature processing), 1 for undifferentiated cells, 2 for moderately differentiated, and 3 for differentiated. 
    \item Once completed, users can use the \verb|pycellmech_nifti| command line to create the NifTI file. For each binary mask file, the function reads the mask, loads the corresponding CSV file, and maps the user-defined final labels. It updates the labeled mask (so users can conduct any quality checks to ensure ROIs are correctly labeled), corrects the orientation of the labeled mask, converts it to a SimpleITK image, and saves it as a NIfTI file. The function then verifies the label consistency between the final labeled mask and the CSV file, ensuring all expected labels are present and identifying any missing or extra labels.
\begin{verbatim}
    pycellmech_nifti 
    --folder_path /path/to/folder/with/binary/masks
    --input_csv_folder /path/to/extract/csv/labels
    --nifti_save_dir /path/to/save/nifti/file
    --label_save_dir /path/to/save/updated/labeled/images
\end{verbatim}
\end{itemize}

The produced NifTI file can then be read into the main \verb|pycellmech| command line to be processed. 

\begin{table}[!t]
\caption{Sample Shape Features Extracted from Binarized Masks in Kvasir Dataset}
\centering
\resizebox{\columnwidth}{!}{%
\begin{tabular}{@{}ccccc@{}}
\toprule
image\_name                   & contour\_number & abe    & mbr\_width & mbr\_angle \\ \midrule
ck2da7fwcjfis07218r1rvm95.jpg & 1               & -0.017 & 180.40     & 53.13      \\
ck2bxskgxxzfv08386xkqtqdy.jpg & 1               & -0.008 & 100.00     & 40.10      \\
ck2bxskgxxzfv08386xkqtqdy.jpg & 2               & -0.050 & 203.93     & 81.87      \\
ck2395w2mb4vu07480otsu6tw.jpg & 1               & -0.104 & 281.34     & 4.86       \\
ck2bxw18mmz1k0725litqq2mc.jpg & 1               & -0.012 & 408.88     & 4.43       \\
ck2bxqz3evvg20794iiyv5v2m.jpg & 1               & -0.020 & 252.97     & 64.26      \\
\end{tabular}%
}
\label{tab:tab1}
\end{table}

\section{Conclusions}\label{sec4}
The objective of \emph{PyCellMech} is to provide users with a one-stop-shop shape feature extractor that is readily available and easy to use. \emph{PyCellMech} can be used alongside existing packages, such as \emph{pyradiomics}, and offers users additional features that may be useful in their medical or biological analyses. The potential use and application of \emph{PyCellMech} can be summarized as follows:

\begin{itemize}
    \item \textbf{Detection:} Help identify certain classes of ROIs or cells based on their shape characteristics.
    \item \textbf{Classification:} Identify shape features that can be used in the automation of classification.
    \item \textbf{Prognosis and Prediction:} Shape features may assist in predicting the likely course of a disease. For example, in a previous study by the authors (\cite{Arslan2023}), shape characteristics played a part in the extraction of different levels of hyperfluorescent regions, which are biomarkers typically observed in the evaluation of the ocular disease geographic atrophy - a cause of irreversible vision loss in individuals aged 50 years and older. 
\end{itemize}

Future iterations of the pipeline will include the addition of other classes of shape features, including scale-space and transform domain. While the results presented are preliminary, the pipeline will be extended to include all shape features over time, and will continually be updated as the research in this space evolves. Furthermore, \emph{PyCellMech} will eventually include automated analyses (e.g., ranking of shape features) and baseline biological interpretations (e.g., certain characteristics could be indicative of healthy or unhealthy cells). In line with the objectives of our laboratory, we will add modules to the pipeline in which shape features will be coupled with mechanobiology to study the implication of shape data on different diseases or cellular states. While the predominant focus of this paper has been on cell-based analyses, such applications can be extended to other medical areas, including the assessment of shape data as related to MRIs. 

\section{Competing interests}
No competing interest is declared.

\section{Author contributions statement}
J.A. developed the idea, created \emph{PyCellMech}, and wrote the manuscript. H.C., I.K., R.V., and K.B. reviewed \emph{PyCellMech} and the manuscript.

\section{Acknowledgments}
The research leading to these results has received funding from the national program “Investissements d’avenir” ANR-10- IAIHU-0006.

\bibliographystyle{unsrt}
\bibliography{pycellmech.bib}

\begin{thebibliography}{1}

\bibitem{vanGriethuysen2017}
Joost~J.M. van Griethuysen, Andriy Fedorov, Chintan Parmar, Ahmed Hosny, Nicole Aucoin, Vivek Narayan, Regina~G.H. Beets-Tan, Jean-Christophe Fillion-Robin, Steve Pieper, and Hugo~J.W.L. Aerts.
\newblock Computational radiomics system to decode the radiographic phenotype.
\newblock {\em Cancer Research}, 77(21):e104–e107, October 2017.

\bibitem{Bookstein1997}
Fred~L. Bookstein.
\newblock Shape and the information in medical images: A decade of the morphometric synthesis.
\newblock {\em Computer Vision and Image Understanding}, 66(2):97–118, May 1997.

\bibitem{Lobo2015}
Joana Lobo, Eugene Yong-Shun See, Manus Biggs, and Abhay Pandit.
\newblock An insight into morphometric descriptors of cell shape that pertain to regenerative medicine: Cell shape analysis descriptors that pertain to regenerative medicine.
\newblock {\em Journal of Tissue Engineering and Regenerative Medicine}, 10(7):539–553, March 2015.

\bibitem{Adams2013}
Dean~C. Adams and Erik Otárola‐Castillo.
\newblock geomorph: an r package for the collection and analysis of geometric morphometric shape data.
\newblock {\em Methods in Ecology and Evolution}, 4(4):393–399, April 2013.

\bibitem{ChakiDey2019}
Jyotismita Chaki and Nilanjan Dey.
\newblock {\em A Beginner's Guide to Image Shape Feature Extraction Techniques}.
\newblock Taylor \& Francis Group, 2019.

\bibitem{Yin2008}
Peng-Yeng Yin.
\newblock {\em Pattern Recognition Techniques, Technology and Applications}.
\newblock InTech, November 2008.

\bibitem{KvasirDataset}
Konstantin Pogorelov, Kristin~Ranheim Randel, Carsten Griwodz, Sigrun~Losada Eskeland, Thomas de~Lange, Dag Johansen, Concetto Spampinato, Duc-Tien Dang-Nguyen, Mathias Lux, Peter~Thelin Schmidt, Michael Riegler, and P{\aa}l Halvorsen.
\newblock Kvasir: A multi-class image dataset for computer aided gastrointestinal disease detection.
\newblock In {\em Proceedings of the 8th ACM on Multimedia Systems Conference}, MMSys'17, pages 164--169, New York, NY, USA, 2017. ACM.

\bibitem{Arslan2023}
Janan Arslan and Kurt Benke.
\newblock Automation of cluster extraction in fundus autofluorescence images of geographic atrophy.
\newblock {\em Applied Biosciences}, 2(3):384–405, July 2023.

\end{thebibliography}

\end{document}